\documentclass[10pt,twocolumn,letterpaper]{article}

\usepackage{cvpr}              

\usepackage{graphicx}
\usepackage{amsmath}
\usepackage{amssymb}
\usepackage{booktabs}
\usepackage{multirow}

\usepackage{times}
\usepackage{epsfig}
\usepackage{graphicx}
\usepackage{amsmath}
\usepackage{amssymb}
\usepackage{bbm}
\usepackage{mathtools}
\usepackage{subcaption}
\usepackage{url} 
\usepackage{sidecap}
\usepackage{pifont}
\usepackage{multirow}
\usepackage[table,xcdraw]{xcolor}
\usepackage{graphicx}
\usepackage{amsmath}
\usepackage{amssymb}
\usepackage{booktabs}
\usepackage{multirow}
\usepackage[ruled, lined, linesnumbered, commentsnumbered, longend]{algorithm2e}

\usepackage{subcaption}

\usepackage{glossaries}
\newacronym{prosa}{ProSA}{Prototype Swapping Autoencoder}
\newacronym{ecr}{ECR}{Elementary Concept Reasoning}
\newacronym{csn}{CSN}{Concept Swapping Network}
\newacronym{icsn}{iCSN}{Interactive Concept Swapping Network}
\newacronym{dt}{DT}{Decision Tree}
\newacronym{lr}{LR}{Logistic Regression}
\newacronym{xil}{XIL}{Explanatory interactive learning}

\usepackage{xcolor}

\usepackage[pagebackref,breaklinks,colorlinks]{hyperref}

\usepackage[capitalize]{cleveref}
\crefname{section}{Sec.}{Secs.}
\Crefname{section}{Section}{Sections}
\Crefname{table}{Table}{Tables}
\crefname{table}{Tab.}{Tabs.}

\def\cvprPaperID{10710} 
\def\confName{CVPR}
\def\confYear{2022}

\begin{document}

\title{Interactive Disentanglement: \\Learning Concepts by Interacting with their Prototype Representations}

\author{Wolfgang Stammer$^{1,3}$
\and 
Marius Memmel$^{1}$
\and
Patrick Schramowski$^{1,3}$
 \and
 Kristian Kersting$^{1,2, 3}$
 \and
 $^{1}$Computer Science Department, TU Darmstadt; 
 $^{2}$Centre for Cognitive Science, TU Darmstadt \\ 
 $^{3}$Hessian Center for AI (hessian.AI) \\
 {\tt\small \{wolfgang.stammer@cs, marius.memmel@stud, schramowski@cs, kersting@cs\}.tu-darmstadt.de}
}

\maketitle

\begin{abstract}
Learning visual concepts from raw images without strong supervision is a challenging task. In this work, we show the advantages of prototype representations for understanding and revising the latent space of neural concept learners. For this purpose, we introduce interactive Concept Swapping Networks (iCSNs), a novel framework for learning concept-grounded representations via weak supervision and implicit prototype representations. iCSNs learn to bind conceptual information to specific prototype slots by swapping the latent representations of paired images. This semantically grounded and discrete latent space facilitates human understanding and human-machine interaction. We support this claim by conducting experiments on our novel data set ``Elementary Concept Reasoning'' (ECR), focusing on visual concepts shared by geometric objects.\footnote{Code available at: \url{https://github.com/ml-research/XIConceptLearning}} 
\end{abstract}

\section{Introduction}



Learning an adequate representation of concepts from raw data without strong supervision is a challenging task. However, it remains important for research in areas of knowledge discovery where sufficient prior knowledge is missing, and the goal is to attain novel understandings. With better representations and architectural components of machine learning models, this appears to become more and more achievable~\cite{Senior0JKSGQZNB20}. However, if remained unchecked this bears the danger of learning incorrect concepts or even confounding features \cite{degrave2021ai, SchramowskiSSA2020}. A further difficult aspect of concept learning, regardless of the level of supervision, is its dynamic and subjective nature. One downstream task might require more fine-grained concepts than others, but also when encountering evidence on novel concepts (\eg in an online learning setting), the knowledge and hierarchy of concepts should be constantly re-approved, discussed, and possibly updated.
It thus remains desirable that the representations learned by such concept learners to be human-understandable and revisable.

\begin{figure}[t]
    \centering
    \includegraphics[width=0.9\linewidth]{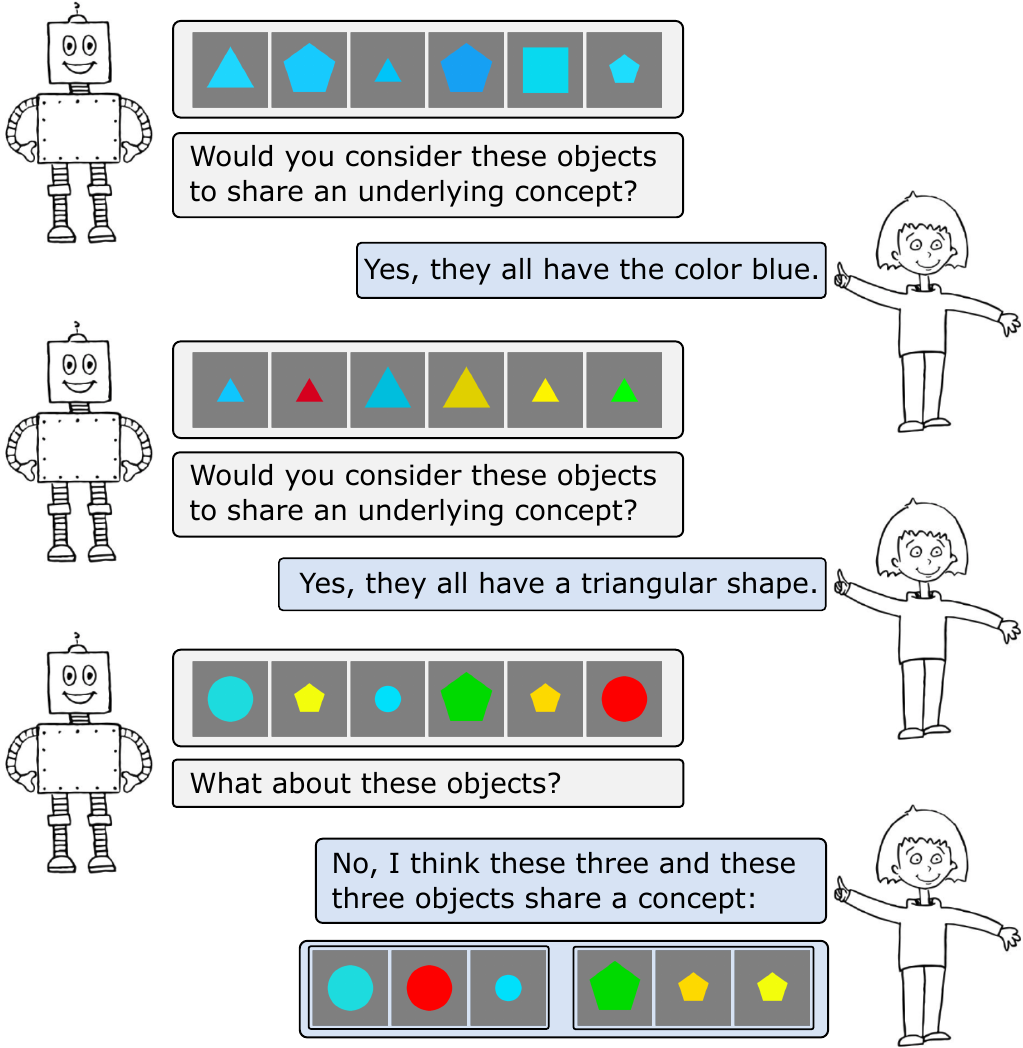}
    \caption{A trained model (left) queries the human user (right) if the concepts that it has extracted from the data coincides with the knowledge of the user. Subsequently, the model can receive revisions from the user.}
    \label{fig:namegame1}
\end{figure}

An evident approach to teaching concept information to a machine learning model is to train it in a supervised fashion through symbolic representations, \eg, one-hot encoding vectors and corresponding raw input~\cite{yi2018neural, KohNTMPKL20}. However, this requires extensive prior knowledge of relevant concepts and seems impractical given the subjective and dynamic nature of concept learning.

\begin{figure}[t]
    \centering
    \includegraphics[width=0.9\linewidth]{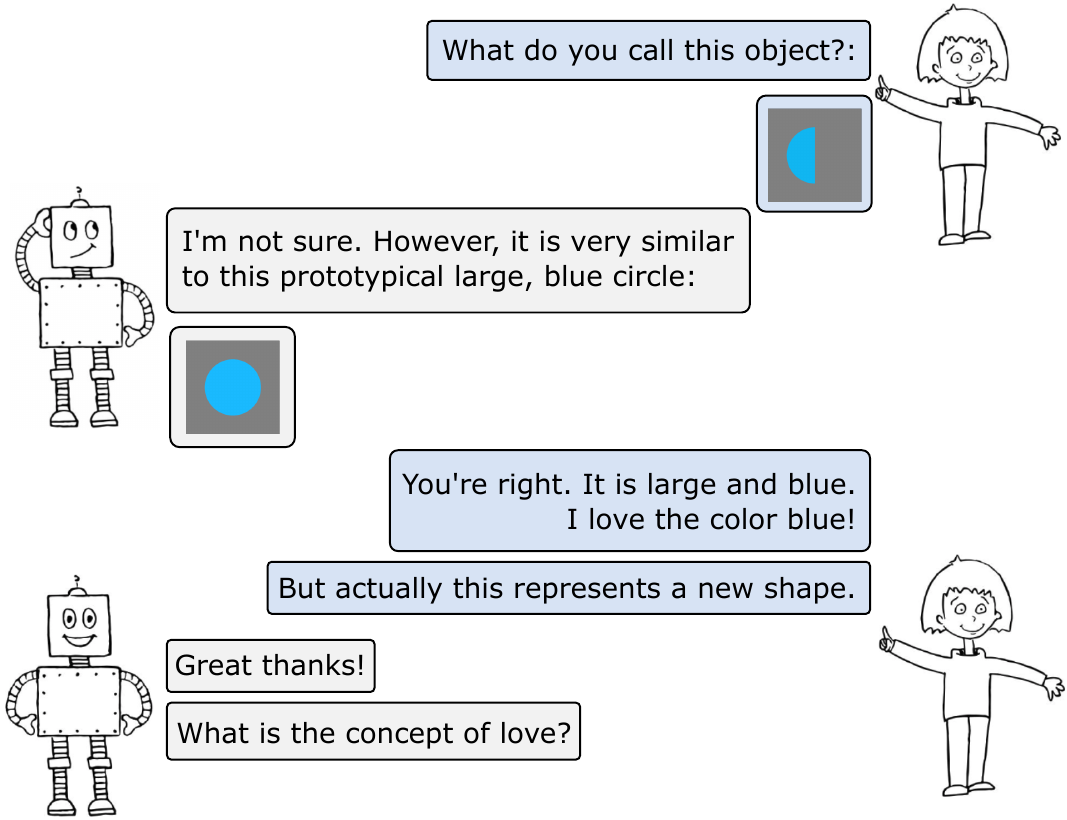}
    \caption{Human-machine interaction for learning about novel concepts. The user queries an object and guides the machine's prototype suggestion if necessary.} \label{fig:namegame2}
\end{figure}

Another branch of research focuses on learning disentangled latent distribution models~\cite{GoodfellowPMX2014, HigginsMPBGBML17, TschannenBL2018}. 
Although initially focused on unsupervised approaches, many recent studies have shifted away from unsupervised learning and show promising results with weak supervision \cite{LocatelloBLRG2019, ShuCKEP20, LocatelloPRSBT20, VCB2020, MeilaZ2021, InsuWMG2021}.
An often implicit assumption of disentanglement research is that the learned latent representations should correspond to human-interpretable factors. Many state-of-the-art variational \cite{LocatelloPRSBT20, MeilaZ2021} and generative adversarial~\cite{Chen2016, NiemeyerG2020, NguyenRMY2020, LiCWW2019, MemmelGM2021} approaches, however, learn continuous latent representations, making these difficult for a human to understand without additional techniques for interpreting the latent space ~\cite{RossCHGD21}. 

Due to the intricate nature of concept learning and inspired by findings on concept prototypes in the fields of psychology and cognitive science, we investigate the advantages of prototype representations in learning human-understandable and revisable concept representations for neural concept learners.
To this end, we introduce the novel framework of \gls{icsn} that learns to implicitly bind semantic concepts to latent prototype representations through weak supervision. This binding is enforced via a discretized distance estimation and swapping of shared concept representations between paired data samples. Among other things, \gls{icsn}s allow for querying and revising its learned concepts \textit{cf}. \cref{fig:namegame1}, and integrating knowledge about unseen concepts \textit{cf}. \cref{fig:namegame2}.

Explicitly focusing on learning object-centric visual concepts, we develop \gls{ecr}, a novel data set containing images of 2D geometrical objects and perform multiple experiments, emphasizing the advantages of our approach.
To summarize, our work highlights the advantages of prototype representations for (i) learning a consistent and human-understandable latent space through weak supervision, (ii) revising concept representations via human interactions, and (iii) updating these in an online learning fashion.

\begin{figure*}[t!]
    \centering
    \includegraphics[width=.95\linewidth]{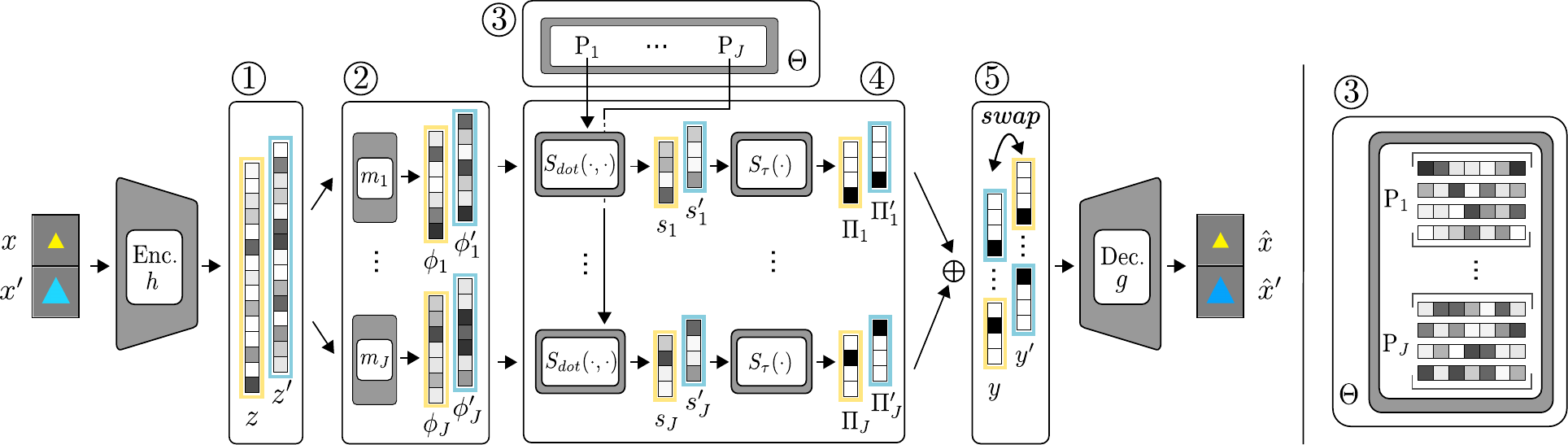}
    \caption{Interactive Concept Swapping Network. An \gls{icsn} is based on a deterministic autoencoder structure providing an initially entangled latent encoding (1). Several read-out encoders (2) extract relevant information from this latent space and compare their extracted concept encodings to a set of prototype slots (3) via a weighted, softmax-based dot product (4). This leads to a discretized code that indicates the most similar prototype slot of each concept encoding. \gls{icsn}s are trained via a simple reconstruction loss, weak supervision via match pairing and a swapping approach that swaps (5) the latent concept representations for shared concepts, enforcing the binding of semantic information to specific prototype representations.
    \label{fig:model}
    }
\end{figure*}

\section{Related Work}
\noindent
\textbf{Concept learning.}
Many previous concept learning approaches focus on predicting selected high-level concepts for improving additional downstream tasks~\cite{yi2018neural, KohNTMPKL20, BelemBSB2021}. 
Several studies highlight the benefits of concept-based machine learning for explainability~\cite{AlvarezMelisJ18, YehKAL2020, GhorbaniWZK2019, MarcosFLF2021} and human interactions~\cite{StammerSK2021}.
To communicate the concepts to a human user, some approaches include first-order logic formulas~\cite{CiravegnaBGGL2021}, causal relationships~\cite{YangLCS2021}, user defined concepts~\cite{KimWGCW2018}, prediction of intermediate dataset-labels~\cite{BelemBSB2021, MarcosFLF2021}, and one-hot encoded bottlenecks~\cite{KohNTMPKL20}. All of these approaches, however, focus on supervised concept learning.
\newline
\textbf{Concept representations in psychology.}
The term concept is rooted in psychology where it can be defined as ``\textit{the label of a set of things that have something in common}''~\cite{archer1966psychological}, though different notions do exist~\cite{fodor1998concepts}.
Most common approaches to represent concepts are exemplars and prototypes. Where the former approach assumes that one or multiple typical examples of a concept are maintained in memory, the latter only assumes an average representation over several observed examples~\cite{Seel2012, OppenheierTK2013, Taylor2003, Smith2014, JaekelSW2008}.
The lines between exemplar and prototype representation become more blurred in the field of cognitive psychology, and their contribution to concept representations is still an open problem, with recent work hinting at the use of both representations~\cite{BowmanIZ2009, FrixioneL2012}.
Nonetheless, there remains evidence of the use and importance of prototypes in the human memory system~\cite{ColcombeW2002, HeindelFOL2013, IbbotsonT2009}. Inspiration sparked by such findings gave rise to Prototype Learning Systems~\cite{Zeithamova2012}.
\newline
\textbf{Neural prototype learning.}
Recent approaches to artificial prototype learning systems focus on neural networks with prototype vectors as internal latent representations. These vectors can be converted into explainable visualizations via decoding approaches~\cite{LiLCR2018} or used for finding the most similar training example~\cite{ChenLTBR2019}. In both works, a class prediction is made based on the similarity of an encoded input to the model's prototypes via a simple distance metric. Lastly, especially in the context of few-shot learning, prototypes show advantageous properties~\cite{SnellSZ17, PrabhudesaiLPTH21, PahdePKN21, HuangZWY2021}.
\newline
\textbf{Disentanglement.}
The field of disentanglement research is also closely related to our work. 
Here the goal is to extract independent underlying factors that are responsible for generating the data~\cite{BengioCV13}.
Recently, through the work of Locatello \etal.~\cite{LocatelloBLRG2019} much of disentanglement research has shifted from unsupervised learning to the weakly supervised setting. Shu \etal.~\cite{ShuCKEP20} show that supervision via match pairing for a known subset of factors gives guarantees for disentanglement via their defined calculus of consistency and restrictiveness.
Literature also extends to group-based disentanglement, allowing for grouping of the identified generative factors~\cite{Hosoya2019, Hosoya2018, ZhuXT2021, BouchacourtTN2018, BillardDPM2018, ZaidiBGC2020}. However, the interpretation of the latent representations from these approaches remains an open question~\cite{RossCHGD21}.
\newline
\textbf{\gls{xil}.} The notion of human interactions on a model's latent concept representations, \eg, to correct confounding behaviour, is closely related to the field of \gls{xil}~\cite{Ross2017right, Selvaraju2019taking, TesoK2019, SchramowskiSSA2020, StammerSK2021}. Specifically, \gls{xil} incorporates the human user into the training loop by allowing for them to interact via a model's explanations. Rather than interacting via post-hoc explanations of previous \gls{xil} approaches, we focus on interacting directly with the latent representations of a model. These, nonetheless, possess a connection to the model's explanations in our setup. Even though we take a more direct approach to revising a model's internal representations, similar feedback methods as in \gls{xil} are applicable for our work. 

\section{\acrlong{icsn}s}
\label{sec:method}
In this section, we explain the basic architectural components of an \acrfull{icsn} before introducing the training procedure and how to interact with the implicit prototype representations of these networks. For an overview, see \cref{fig:model}.

\noindent\textbf{Prototype-based concept architecture}. Assume an input $x_i \in X$, with \mbox{$X := [x_1, ..., x_N] \in \mathbb{R}^{N \times D}$}. For the sake of simplicity, we remove the sample index $i$ from further notations below and denote with $x\in\mathbb{R}^{D}$ an entire image. However, in our framework, $x$ can also be a latent representation of a subregion of the image. This subregion can be implicitly or explicitly extracted from the image by a pre-processing step, \eg via segmentation algorithms~\cite{Girshick15, he2017mask, RedmonDGF16, CarionMSUKZ20} or compositional generative scene models~\cite{burgess2019monet, engelckeGenesis, GreffIODINE, LinSPACE, SlotAttention, Du2020CompositionalVG, du2021unsupervised}. Additionally, we assume each $x$ to contain several attributes such as color, shape and size. Specifically, we refer to the realizations of these attributes, \eg, a ``blue color'' or ``triangular shape'' as a \textit{basic concept}. In contrast, we refer to ``color''  as a category concept or, as often called in the field of cognitive and psychological sciences, \textit{superordinate concept}~\cite{eysenck2018fundamentals}. Each image $x$ therefore has the ground truth basic concepts $c := [c_1, ..., c_J]$ with $J$ denoting the total number of superordinate concepts. We make the necessary assumption that $x$ can only contain one basic concept realization per superordinate concept. 
For simplicity, we furthermore assume that each superordinate concept contains the same number of basic concepts $K$ which might vary in practice as we are going to show in our experiments.

Assuming an encoder-decoder structure, we define an input encoder $h(\cdot)$ that receives the image $x$ and encodes it into a latent representation $z \in \mathbb{R}^{Z}$ by $h(x) = z$. Rather than reconstructing directly from $z$, as done by many autoencoder-based approaches, an \gls{icsn} first applies several read-out encoders $m_j(\cdot)$ to the latent representation $z$ resulting in $m_j(z) = \phi_j \in \mathbb{R}^{Q}$ with $j \in [1, ..., J]$. We refer to the encoding $\phi_j$ as a \textit{concept encoding}. The goal of each read-out encoder is to extract the relevant information from the entangled latent space $z$ that corresponds to a superordinate concept, \eg color. 
We discuss how we enforce this extraction of concept specific information below. 

One central component of the \gls{icsn} is a set of codebooks each containing multiple prototype slots. We define this set as $\Theta := [\mathrm{P}_1, ..., \mathrm{P}_J]$, with a single codebook as $\mathrm{P}_j \in \mathbb{R}^{Q \times K}$. Each codebook contains an ordered set of trainable, randomly initialized prototype slots $p_{j}\in\mathbb{R}^{Q}$, i.e., $\mathrm{P}_j := [p_{j}^{1}, ... , p_{j}^{K}]$. 

To enforce the assignment of each concept encoding $\phi_j$ to one prototype slot of $\mathrm{P}_j$, we define a similarity score $S_{dot}(\cdot,\cdot)$ as a softmax over the dot product between its two inputs. This way we obtain the similarity between a concept encoding, $\phi_j$, and specific prototype slot, $p^{k}_{j}$ with:
\begin{align}
    s^{k}_{j} = S_{dot}(\phi_j, p^{k}_{j}) = \frac{\exp{(\phi_j \cdot p^{k}_{j} / \sqrt{Q})}}{\sum_{k=1}^{K} \exp{(\phi_j \cdot p^{k}_{j} / \sqrt{Q})}}
\end{align}
The resulting similarity vector $s_j\in\mathbb{R}^{K}$ contains the similarity score for each prototype slot of category $j$ with the concept encoding $\phi_j$. To enforce further discretization and the binding of concepts to individual prototype slots, we introduce a second function $S_{\tau}(\cdot)$ to apply a weighted softmax function to the similarity scores: 
\begin{align}
    \Pi^{k}_{j} = S_{\tau}(s^{k}_{j}) = \frac{\exp{(s^{k}_{j}/\tau)}}{\sum_{k=1}^K\exp{(s^{k}_{j}/\tau)}}, 
\end{align}
with $\Pi_j \in \mathbb{R}^{K}$ and weight parameter $\tau \in \mathbb{R^+}$. In our experiments we step-wise decrease $\tau$ to gradually enforce the binding of information. In the extreme case of $\tau\to 0$, $\Pi_j$ resembles a one-hot vector (multi-label one-hot vector in the case of $J > 1$), indicating which prototype slot of category $j$ the concept encoding $\phi_j$ is most similar to. 

Finally, we concatenate the weighted similarity scores of each category into a single vector to receive the final prototype distance codes $y := [\Pi_{1}, ..., \Pi_{J}] \in [0, 1]^{J\cdot K}$ which we pass to the decoder $g(\cdot)$ to reconstruct the image: $g(y) = \hat{x} \in \mathbb{R}^{D}$. 



\noindent \textbf{Concept swapping and weak supervision}. Prior to training, i.e., after initialization, there is no semantic knowledge bound to the prototype slots yet. Each prototype carries just as little meaning as the other. The semantic knowledge found in converged \gls{icsn}s, however, is indirectly learned via a weakly-supervised training procedure and by employing a simple swapping trick. 

We adopt the match pairing approach of Shu \etal.~\cite{ShuCKEP20}, a practical weakly-supervised training procedure to overcome the issues of unsupervised disentanglement~\cite{LocatelloBLRGSB19}. In this approach, a pair of images $(x, x')$ is observed that shares values for a known subset of underlying factors of variation within the data, \eg color, while the total number of shared factors can vary between $1$ and $J-1$. 
In this way, a model can use the additional information of the pairing to constrain and guide the learning of its latent representations. 

Previous works on weakly-supervised training, specifically of VAEs, reverted to applying a product~\cite{BouchacourtTN2018} or an average~\cite{Hosoya19} of the encoder distributions of $x$ and $x'$ at the shared factor IDs. Locatello \etal.~\cite{LocatelloPRSBT20}, extended these works to a setting with an even weaker form of supervision but carries fewer disentanglement guarantees. 
In comparison to these works, an \gls{icsn} uses a simple swapping trick between paired representations, similar to Caron \etal.~\cite{CaronMMGBJ20}. 
Specifically, with $v$ being the shared factor ID between the image pairs $(x, x')$ the corresponding similarity scores $(\Pi_{v},\Pi'_{v})$ are swapped between the final corresponding prototype codes, resulting in: 
\begin{align*}
    y := [\Pi_{1}, ..., \mathbf{\Pi'_{v}}, ..., \Pi_{J}], \ 
    y' := [\Pi'_{1}, ..., \mathbf{\Pi_{v}}, ..., \Pi'_{J}].
\end{align*}
This swapping procedure has the intuitive semantic that it forces an \gls{icsn} to extract information from the first image that it can use to represent properties of the category $v$ of the second image. 

Pseudo-code of \gls{icsn}s can be found in the Supplementary Materials.

\noindent \textbf{Training objective}. \gls{icsn}s are finally trained with a single pixel-wise reconstruction loss per paired image over batches of size $N$: 
\begin{equation}\label{eq:loss}
    L = \frac{1}{2N}\sum\nolimits_{i=1}^{N}{(x_i - \hat{x}_i)^2 + (x'_i - \hat{x}'_i)^2}
\end{equation}

This simple loss term stands in contrast to several previous works on prototype learning, which enforce semantic binding via an additional consistency loss~\cite{LiLCR2018, PrabhudesaiLPTH21, ZMichieliZ21}. 
By including the semantic binding implicitly into the network architecture, we eliminate the need for additional hyperparameters and a more complex optimization process over multiple objectives. 

\noindent \textbf{Interacting with iCSNs}. The goal of \gls{icsn}s, especially in comparison to VAEs, is not necessarily to be a generative latent-variable model that learns the underlying data distribution, but to learn prototypical concept representations that are human-understandable and interactable. The autoencoder structure is thus a means to an end rather than a necessity. However, instead of discarding the decoder after convergence, an \gls{icsn} can present an input sample's closest prototypical reconstruction of each concept. Thus, by querying these prototypical reconstructions at test time, a human user can confirm whether the predicted concepts make sense and possibly detect undesired model behavior. By defining a threshold on the test time reconstruction error, an \gls{icsn} can give a heuristic indication of its certainty in recognizing concepts in novel samples.

Due to the discrete and semantically bound latent code $y$, a human user can easily interact with \gls{icsn}s by treating $y$ as a multi-label one-hot encoding. Specifically, a human user can revise and add knowledge via additional loss terms to the full extent of Stammer \etal.~\cite{StammerSK2021}. For example, via logical statements such as $\forall img. \Rightarrow \neg hasconcept(img, p^1_1)$ or $\forall img.\ isin(img, imgset) \Rightarrow hasconcept(img, p^1_2)$, a user can formulate logical constraints which read as ``Never detect the concept represented by the prototype $p^1_1$.'' and ``For every image in this set of images you should be detecting the concept represented by prototype $p^1_2$.'', respectively. The set of incorrectly represented images can be curated by the user interactively.

Lastly, the modularity of \gls{icsn}s has additional advantages or interactive online learning, \eg, when the model is provided with data samples that contain novel concepts or when a factor that is present in the data is initially deemed unimportant but considered important retrospective to the initial learning phase. The approach for interaction in both cases depends on the hierarchy of the concept to be learned, namely if it is a basic concept or a superordinate concept. In the case of a novel basic concept, the approach is straightforward. Assuming human user satisfaction with the previous concept representations of an \gls{icsn}, and that $J$, the total number of prototype slots per codebook, was set to be overestimated, a user can simply give the feedback to represent a novel basic concept via one of the unused prototype slots of the relevant category.

In case a novel superordinate concept should be learned, one can apply a simple trick during the initial training phase by adding an additional read-out encoder $\tilde{m}_{J+1}(z) = \tilde{\phi}_{J+1} \in \mathbb{R}^{L}$. In contrast to the other read-out encoders, this one does not map to the space of the prototype slots where it would consecutively be discretized via $S_{dot}(\cdot,\cdot)$ and $S_{\tau}(\cdot)$. Instead, $\tilde{\phi}_{J+1}$ remains a continuous representation that is directly concatenated to the final latent codes $y := [\Pi_{0}, \Pi_{1}, ..., \Pi_{J}, \tilde{\phi}_{J+1}]$. In this way, $\tilde{m}_{J+1}(\cdot)$ can learn to incorporate all information that should not be discretized via the usual procedures. Ultimately, the initial latent space $z$ of an \gls{icsn} can be trained to represent the full data distribution, even though only specific concepts should be discretized from this space. To include concepts that were initially considered irrelevant, one can just expand $J$ which means to add a new read-out encoder $m_{J+1}(z) = \phi_{J+1} \in \mathbb{R}^{Q}$ and codebook $\mathrm{P}_{J+1}$ to the \gls{icsn}. Then, $m_{J+1}$ learns to bind novel basic concepts from the ``novel'' superordinate concept to $\mathrm{P}_{J+1}$ which only requires novel data pairs exemplifying the previously unimportant concept. 

\noindent \textbf{Additional remarks}. To summarize: the gradient of \cref{eq:loss} provides the learning signal for the entire network, including the initial encoder, read-out encoders, prototype slots and decoder. Furthermore, by decreasing the softmax temperature $\tau$ in a step-wise fashion, one can enforce the binding of information to specific prototypes such that a specific concept is mapped to an individual prototype slot. Through this discretization process, the decoder learns to produce reconstructions that correspond to the prototypical representations present in the data. For example, given a pair of images of blue objects that vary in their shade of blue, an \gls{icsn} would learn to map the color information of these objects to the same prototype slot, thus learning the prototypical blue of both shades. 
This discretization step is a key difference to the various GAN, and VAE approaches with Gaussian distributions, which try to learn a continuous latent space of the underlying factors. 
\section{Elementary Concept Reasoning Data Set}

\begin{figure}[t]
    \centering
    \includegraphics[width=0.9\columnwidth]{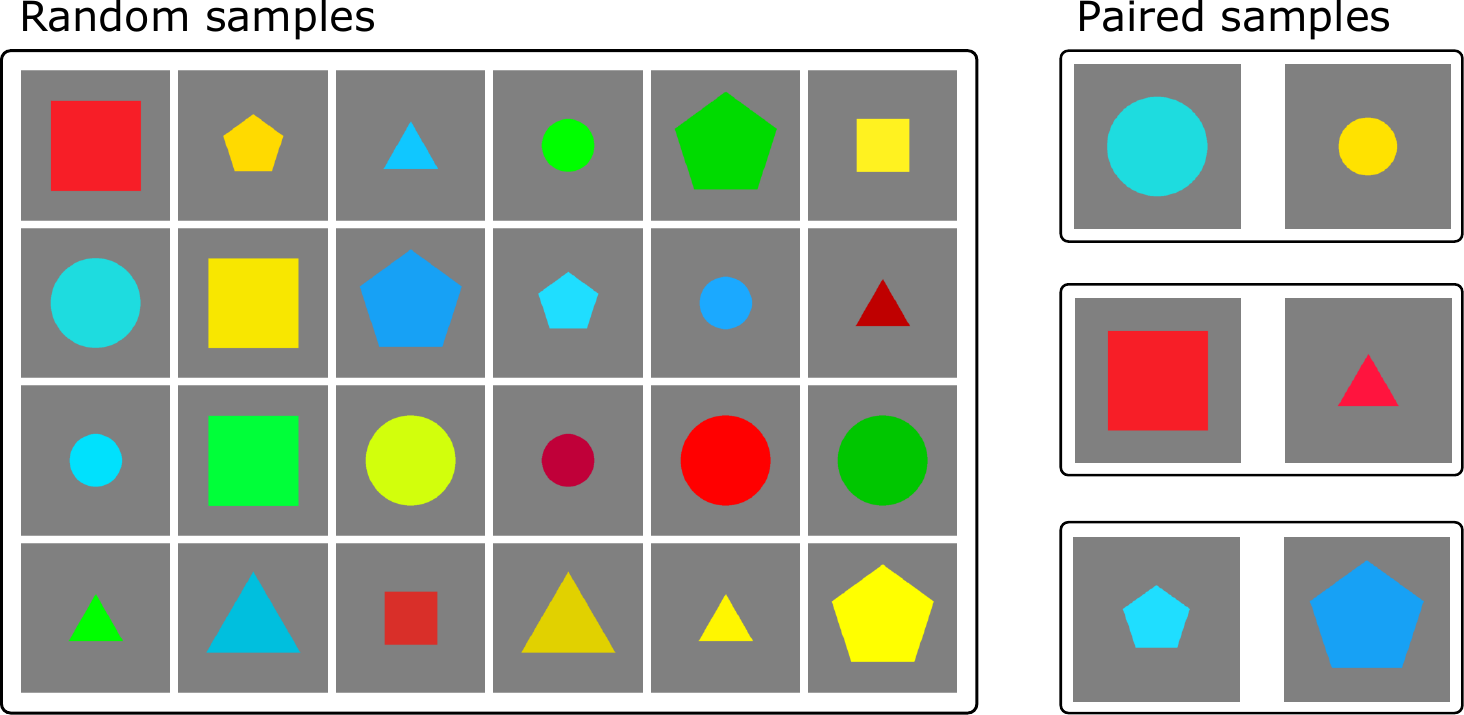}
    \caption{Samples of the \acrlong{ecr} data set. Each sample image (left) depicts a centered 2D object with three different properties: color, shape, and size. Images are paired such that the objects share between one and two concepts (right).}
    \label{fig:dataset}
\end{figure}

Recent studies show the benefits of object-centric learning for performing complex downstream tasks~\cite{yi2018neural, SlotAttention, StammerSK2021}. Thus, rather than learning concepts of an entire image, \eg as Kim \etal \cite{KimM18}, we introduce a novel benchmark data set, \acrfull{ecr}, which explicitly focuses on object-centric visual concept learning. 

\gls{ecr} consists of RGB images ($64\times64\times 3$) of 2D geometric objects on a constant colored background. Objects can vary in shape (circle, triangle, square, and pentagon), size (large and small), and color (red, green, blue, yellow). We add uniform jitter to each color, resulting in various color shades. Each image contains a single object which is fixed to the center of the image.
Furthermore, \gls{ecr} contains image pairs following the match pairing setup of Shu \etal~\cite{ShuCKEP20}. We pair images such that the objects in the individual images share at least one, but at most $J-1$ common properties.
\gls{ecr} contains a training set size of 5000 image pairs and 2000 images for validation.

\Cref{fig:dataset} shows example images of \gls{ecr}. Random samples are presented on the left, exemplifying shape, size, and color combinations. Example image pairs are presented on the right. An important feature of \gls{ecr} is that although various shades of colors exist, they all map to four discrete colors. Notice, \eg, the color difference of the two paired blue objects. Even though both objects present different shades of blue, their state of being paired indicates that they share the same distinct shape (pentagon) and color (blue) concept. 

\section{Results}

In this section, we demonstrate the advantages of prototype-based representations via \acrlong{icsn}s. 
We begin our analysis by investigating the sparsity and semantics of \gls{icsn}s' latent representations. Next, we show that the model can communicate the extracted concepts to a human user due to its discretized latent space. Subsequently, we simulate human user interactions via simple feedback rules, which are sufficient to revise an \gls{icsn}s' latent concept space. Lastly, we show that novel concepts can easily be added into the concept space of \gls{icsn}s via simple human interactions.

\noindent\textbf{Experimental details}. For our experiments, we compare the \gls{icsn} to several baselines including the unsupervisedly-trained \textit{$\beta$-VAE} \cite{HigginsMPBGBML17} and \textit{Ada-VAE} by Locatello \etal.~\cite{LocatelloPRSBT20}, using the arithmetic mean of the encoder distributions as in~\cite{Hosoya19}. For a fair comparison with \gls{icsn}s which are trained via the shared match pairing of \cite{ShuCKEP20} and the Ada-VAE, which was originally introduced as a weaker form of supervision, we also trained the Ada-VAE with known shared factor IDs. This baseline essentially resembles a $\beta$-VAE with an averaging of encoder distributions between pairs of images at the known shared factor IDs. It is denoted as \textit{VAE} in the results below. Lastly, we compare to a discretizing VAE approach which uses a categorical distribution via the Gumbel-softmax trick~\cite{JangGP17, MaddisonMT17} (\textit{Cat-VAE}). Cat-VAE is trained the same way as the VAE, i.e., via share pairing and averaging over encoder distributions. 

We train the \gls{icsn} with a simple reconstruction loss as in \cref{eq:loss} and decreasing softmax temperature. We present the results of two \gls{icsn} configurations. The vanilla setting, denoted as \textit{\gls{csn}}, corresponds to an \gls{icsn} prior to human interactions with the correct number of superordinate concepts ($J=3$) and an overestimated number $K$ of prototype slots per superordinate concept ($K=6$). 
Finally, \textit{\gls{icsn}} denotes a \gls{csn} after the initial training phase with additional user interactions.

The number of latent variables for \textit{$\beta$-VAE}, \textit{Ada-VAE}, and \textit{VAE} was set to the ground truth number of superordinate concepts.
All \textit{Cat-VAE} runs were performed with three categorical distributions each with $k=6$ events. 

All configurations were trained with five random seed initializations, and the results present the mean and standard deviation of these runs. All presented results were obtained from a held-out validation set. Further details can be found in the Supplementary Materials.

\begin{figure}[t]
    \centering
    \includegraphics[width=0.98\columnwidth]{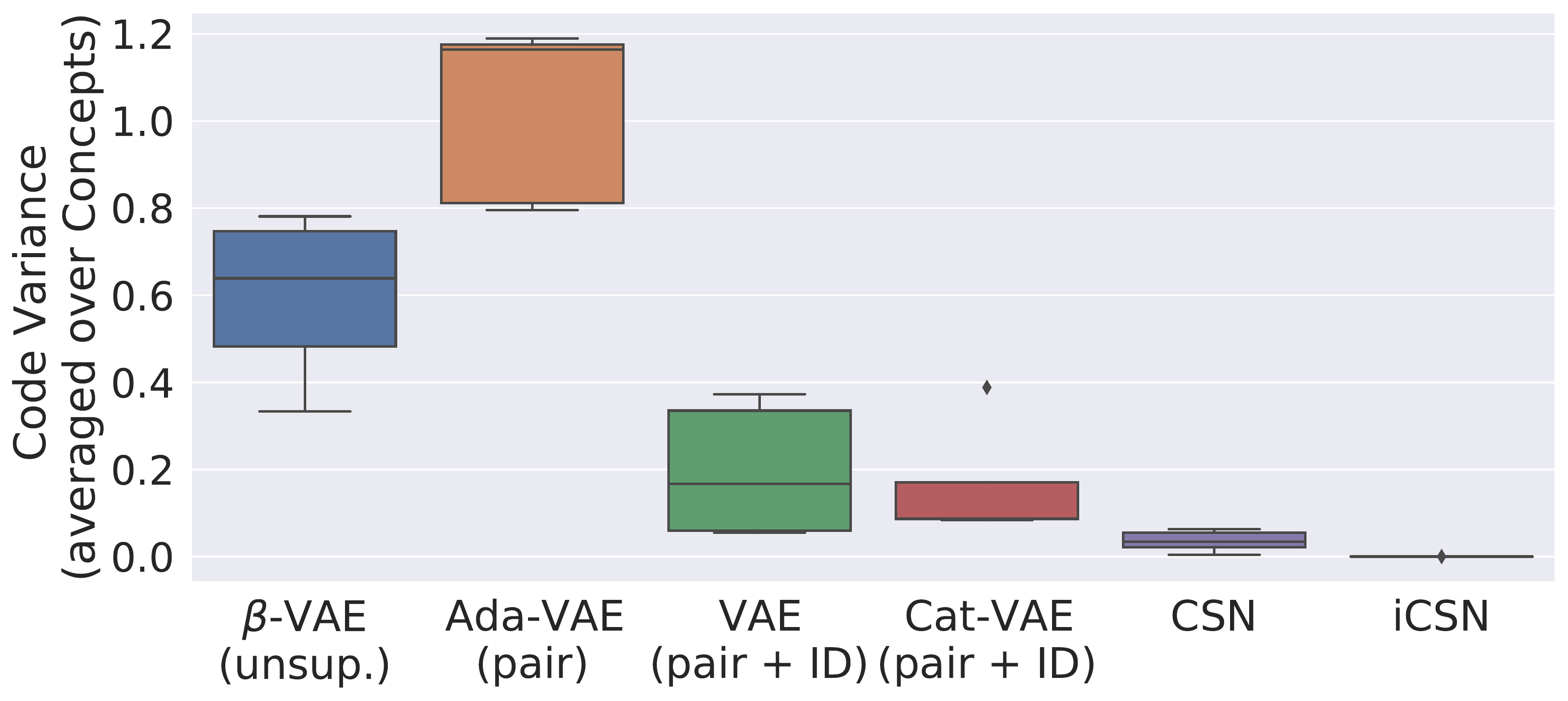}
    \caption{Average latent code variance given the ground truth concept labels for different model types and training settings. The compared models: unsupervised trained $\beta$-VAE, Ada-VAE with paired images, VAE and categorical VAE with paired images and known shared factor IDs, the novel \gls{csn} and \gls{icsn} with additional interactions on the learned concept space. Note that a lower variance is desirable.  \label{fig:code_var}
    }
\end{figure}

\noindent\textbf{Reduced code variance}. Human understandability of and interactions with a model's latent space strongly benefit from consistency in a model's concept representations. In other words, the representation of the color blue should be specific to this concept and the latent representation for the blue color of one object should be very similar to that of a second blue object. If this is not the case, it remains difficult for a human user to identify and interact with these learned concepts. 

Motivated by this intuition, we first investigate the variance of the latent representations given the ground truth multi-label information of each validation image.
For this, we compute the latent code variance over all validation images of each concept. 
In mathematical notation this corresponds to:
\begin{equation}
    \frac{1}{K\cdot J}\sum_{j=1}^{J=3} \sum_{k=1}^{K} \text{Var}(\{\bar{z}_j\}_{l==k})
\end{equation}
with $\text{Var}(\cdot)$ denoting the variance, $\bar{z}_j$ denoting a place holder for the latent representation of a corresponding model (the discretized prototype distance code $y$ for \gls{icsn}s, the distribution means for VAEs with Gaussian distributions and the event probabilities for Cat-VAE). $\{\cdot\}_{l==k}$ denotes the set of latent representations from images for which the ground truth basic concept of category $j$ corresponds to $k$. 

\begin{table}[t!]
    \small
    \centering
    \begin{tabular}{l l l}
        \multicolumn{1}{l}{}&\multicolumn{1}{c}{DT}&\multicolumn{1}{c}{LR} \\
        \midrule
        $\beta$-VAE (unsup.) & $88.98 \pm 11.53$ & $73.86 \pm 18.07$ \\
        Ada-VAE (pair) & $ 63.39 \pm 11.32$ & $74.07 \pm 6.40$ \\
        VAE (pair + ID) & $97.28 \pm 5.32$ & $97.88 \pm 0.73$ \\ 
        Cat-VAE (pair + ID) & $74.79 \pm 19.45$ & $91.17 \pm 13.13$ \\
        \textbf{\gls{csn}} & $\mathbf{99.92 \pm 0.05}$ & $\mathbf{99.87 \pm 0.07}$ \\
        \textbf{\gls{icsn}} & $\mathbf{100.0 \pm 0.00}$ & $\mathbf{100.0 \ \pm 0.00}$ \\
        \midrule
        \multicolumn{3}{c}{\textbf{Ablation}}\\
        Cat-VAE w. swap. & $60.85 \pm 7.29$ & $65.71 \pm 10.66$ \\
        \textbf{\gls{icsn}} w. avg. & $68.84 \pm 39.4$ & $68.81 \pm 39.4$ \\
    \end{tabular}
    \caption{Linear probing via decision tree (DT) and logistic regression (LR). (top) Probing on the latent codes of iCSN models and various baselines. (bottom) Ablation study via probing on the latent codes of Cat-VAE with encoder distribution swapping and \gls{icsn} concept encoding averaging. 
    All classification accuracies were computed on a held out test set. \label{tab_lin_probe}}
\end{table}

The resulting code variances over all models can be seen in \cref{fig:code_var}. Note that a low code variance is desirable and indicates how well a concept is mapped to a distinct representation. The results in \cref{fig:code_var} suggest that the variance of the latent space from \gls{csn}s is much lower, showing more consistent concept representations.
However, a reduced latent code variance is not a sufficient criterion for concept consistency and human understandability. For example, a model that learns to map all concepts to a single representation has zero latent code variance but also no representational power. Therefore, we turn to probing the latent concept space via linear models next.

\noindent \textbf{Probing the latent space.} Similar to works of the self-supervision community~\cite{CaronMMG2020, PatacchiolaS2020, ChenH2021, ChenKNH2020, ChenXH2021}, we investigate the latent code of each model via linear probing. 
For this, the latent codes of each model on a held-out data set are inferred, as in the previous experiment. Next, ground truth labels are obtained by converting each multi-label ground truth vector, $c$, of this data set to a 32-dimensional one-hot encoding. Finally, a \gls{dt} and a \gls{lr} are trained supervisedly on this data set and validated on an additional held-out data set. 

The results in \cref{tab_lin_probe} (top) document the average accuracy and standard deviation on the held-out validation set over the five random initializations for the different models. 
We observe that the latent code of \gls{csn}s allows for nearly perfect predictive performance and surpasses all variational approaches. Importantly, \gls{csn}s' representations even surpass those of VAE approaches (\textit{VAE} and \textit{Cat-VAE}) that were trained with the same type of weak supervision as \gls{csn}s. As expected, the $\beta$-VAE performs worse on average than the weakly-supervised models. Interestingly, however, the Ada-VAE configuration performed worse than the $\beta$-VAE. In addition, the discrete latent representations of Cat-VAE also perform worse than \gls{csn}s. Noticeably, the Cat-VAE runs indicate a high deviation in performance, indicating that several Cat-VAE runs converged to sub-optimal states.
In summary, although the \gls{ecr} data set only contains variations in individual 2D geometrical objects, the baseline models do not perform as well as \gls{csn}s, even when trained with the same amount of information.




\noindent \textbf{Explaining and revising the latent space.} An advantage of an \gls{icsn}'s semantically bound, discrete latent space, is the straightforward identification of sub-optimal concept representations by a human user \textit{cf.}~\cref{fig:namegame1}. Upon identifying correctly or falsely learned concepts, a user can then apply simple logical feedback rules on this discrete concept space. 

Specifically, after training via weak supervision, it is recommendable for the machine and human user to discuss the learned concepts and identify whether these coincide with the user's knowledge or if a revision is necessary. 
For example, an \gls{icsn} can learn to represent a color over several prototype slots or represent two shapes via one slot, indicating that it falsely considers these to belong to the same concept. 
An \gls{icsn} can then convey its learned concepts in two ways. First, it can group novel images that share a concept according to its inferred discrete prototype distance codes and inquire a human user if indeed the grouped images share a common underlying concept \textit{cf.}~\cref{fig:namegame1}. 
Second, utilizing the decoder, it can present the prototypical reconstruction of each learned concept, \eg, presenting an object with a prototypical shade of blue \textit{cf.}~\cref{fig:namegame2}.


Having identified potential sub-optimal concept representations, a human user can now interact on the discretized latent space of \gls{icsn}s
via logical rules and further improve the representations, which we demonstrate via simulated user interactions in the following. 
For all previous runs of the vanilla \gls{csn} configuration, 
we visually inspect the concept encodings $y$ for one example each of the 32 possible concept combinations and identify those prototype slots which are ``activated'' in the majority of examples per individual concept (\textit{primary} slots) and, additionally, identify those prototype slots per concept that are never or rarely activated within our subset of examples (\textit{secondary} slots). We next apply an \textit{L2} loss on $y$ to never use these \textit{secondary} slots and finetune the previous runs on the original training set with the original reconstruction loss and this additional \textit{L2} loss. The semantics of this feedback is that concepts should only be represented by their \textit{primary} prototype slots. Additionally, in two runs we revise an observed sub-optimal solution that pentagons and circles are bound to the same prototype slot. Hereby, feedback is provided on all pentagon samples of the training set to bind to an otherwise empty prototype slot, again via an additional \textit{L2} loss. 

The results of these interactions can be found under \textit{iCSN} in \cref{fig:code_var} and \cref{tab_lin_probe} (top) indicating a near-zero latent code variance per ground truth concept and perfect linear probing accuracy, respectively. Thus, indicating the ease of interacting with and revising the latent space of \gls{icsn}s. 


\noindent \textbf{Interactively learning novel basic concept.}
Furthermore, the prototype-based representations of \gls{icsn}s possess interesting properties for an online learning setting, \eg, when encountering novel concepts which the model has not seen before. Through the decoder of the \gls{icsn}, evaluating the reconstruction and reconstruction error can serve as a means for identifying whether the model has a good representation of a novel sample \textit{cf.}~\cref{fig:namegame2}. 

\begin{figure}[t!]
    \begin{minipage}[t][][b]{0.22\columnwidth}
    \centering
        \includegraphics[width=0.7\columnwidth]{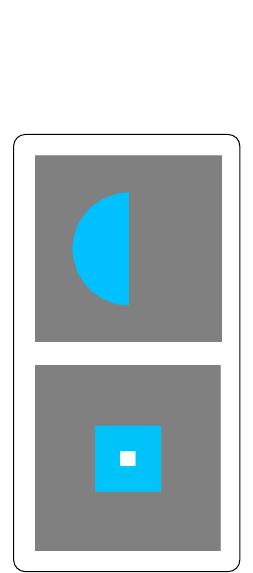}
    \end{minipage}
    \begin{minipage}[t][][b]{0.25\columnwidth}
        \small
        \centering
        \begin{tabular}{l l l}
            \multicolumn{1}{l}{\textbf{iCSN}} & \multicolumn{1}{c}{DT} & \multicolumn{1}{c}{LR} \\
            \midrule
            \multicolumn{3}{c}{Novel Basic Concept}\\
            before & $78.87 \pm 1.31$ & $78.73 \pm 1.22$ \\ 
            after & $\mathbf{98.3 \pm 1.32}$ & $\mathbf{98.28 \pm 1.33}$ \\  
            \midrule
            \multicolumn{3}{c}{Novel Superordinate Concept}\\
            before & $93.1 \pm 4.46$ & $67.54 \pm 9.07$ \\ after & $\mathbf{99.85 \pm 0.3}$ & $\mathbf{98.29 \pm 3.42}$ \\
        \end{tabular}
    \end{minipage}
    \caption{Linear probing via \acrfull{dt} and \acrfull{lr} on the latent codes of \gls{icsn}. We evaluate the models with a \gls{ecr} data set containing a novel shape (top) and a novel superordinate spot concept (bottom), each not seen during initial training (before). After human user interactions (after) this novel information could be easily added to the concept representations. \label{lin_probe_new_shape}}
\end{figure}

When teaching an \gls{icsn} a novel basic concept like a \textit{halfcircle} shape \textit{cf.}~\cref{lin_probe_new_shape}, a user can identify an unbound prototype slot of the model's latent representation and encourage the binding to this slot. To prevent catastrophic forgetting~\cite{McCloskeyN1989,Ratcliff1990}, i.e., overriding already learned concepts, we employ a combination of common rehearsal~\cite{Robins1993} and knowledge distillation methods~\cite{Hinton2015,ZhizhongDBJ2016} by letting the model predict the latent codes of past samples and restrict the \gls{icsn} to not deviate from those. Specifically, we use a simple \textit{L2} loss on the known and unknown one-hot concept encodings to encourage binding of the unknown concepts while not forgetting the known ones.

\Cref{lin_probe_new_shape} (top) shows the linear model prediction accuracies on the latent space of \gls{icsn}s that have been presented a data set containing the novel concept \textit{halfcircle}. The tag \textit{before} indicates the accuracy on the latent code of the \textit{iCSN} runs of \cref{tab_lin_probe} (top) that were trained on the standard \gls{ecr} concepts, whereas \textit{after} indicates the accuracy on the latent code after additional interactions with a simulated user by providing the information which empty prototype slot of the shape codebook to bind the novel concept to. These results indicate the ease of adding additional knowledge on novel basic concepts to the latent representation of \gls{icsn}s.


\noindent \textbf{Interactively learning novel superordinate concept.} Next, we showcase how to add a novel superordinate concept. For this setting, we make use of a variation of the \gls{ecr} data set where white spots were added to the center of roughly half of the objects \textit{cf}.~\cref{lin_probe_new_shape}. The other half depicts a solid color as in the original \gls{ecr} data set. 

We consider an online learning setting where spots are unimportant during the initial training but reconsidered important in the second round of interaction. The modularity of \gls{icsn}s allows us to easily add a read-out encoder during initial training that learns to represent all the information of the data that is not discretized via the initial paired samples (\textit{cf}.~\cref{sec:method} for details). In the second round of training, this continuous read-out encoder can be replaced with a discretizing read-out encoder and additional codebook. This modularity property further eliminates the danger of catastrophic forgetting in that all previously trained modules can be frozen in the second training round, thus only requiring the novel read-out encoder to be finetuned. 

Human user interactions are simulated by assuming that the \gls{icsn} has correctly learned the previous concepts with an additional continuous read-out encoder for representing the spotted feature. Subsequently, additional training pairs are introduced that exemplify the novel superordinate concept, and the new read-out encoder is finetuned as in the standard training setting. Results can be seen in \cref{lin_probe_new_shape} again presenting the linear model accuracy on the latent representations before and after simulated user interactions, indicating that the novel superordinate concept can easily be bound to the model's internal prototype representations.

Here, we remark on desirable properties of \gls{icsn}s for handling confounded data. Assuming an undesired confounding factor within the data generation process that causes spurious features, an \gls{icsn} can learn to ignore these features during its training process via the mechanism presented above. This stands in comparison to GANs or variational approaches with Gaussian distributions, which could potentially learn also to model the spurious features.


\noindent \textbf{Ablation studies.} To assess the importance of the different components of \gls{icsn}s, we conduct an ablation study and depict the linear probing classification performances in \cref{tab_lin_probe} (bottom). Specifically, we test a Cat-VAE that uses swapping of the relevant encoder distributions, rather than averaging as in previous experiments. And, secondly, we test an \gls{icsn} with averaging of the concept encodings $\phi_j$, rather than swapping.
With these experiments, we wish to (i) compare the discretization via distances to prototypes to discretization via categorical distributions (Cat-VAE w. swapping) and (ii) test the influence of swapping versus averaging of encodings for \gls{icsn}s (\gls{icsn} w. avg). 

The results of \cref{tab_lin_probe} (bottom) when compared with \cref{tab_lin_probe} (top) indicate that discrete representations via distances to prototypes are, in fact, beneficial compared to those of the categorical distributions of Cat-VAEs. Secondly, the swapping procedure appears to be crucial for optimal learning of concept representations in \gls{icsn}s.

\noindent \textbf{Limitations.} Following assumptions were made in this work: a superordinate concept is divisible into multiple basic concepts and ``valid'' user feedback was provided in our experiments on interactions. A potential limiting factor of \gls{icsn}s is the training reconstruction loss which might be insufficient for learning fine-grained concepts. 
Additionally, we observed that the choice of $\tau$ can influence the quality of the learning process. Setting a small value too early can lead to sub-optimal solutions.
Lastly, our approach was tested on \gls{ecr} due to its object-centric nature and distinct concept distribution of the data set. For more complex settings 
additional architectures may be required to pre-process the data to \eg extract objects from an image. 

\section{Conclusion}
In this work, we investigated the properties of latent prototype representations for neural concept learning with weak supervision. The results with our novel \gls{icsn} framework indicate that these are beneficial for human-understandable concept learning but also human interactions and the incorporation of novel concepts within an online learning setting. 
Interesting pathways for future research are applying \gls{icsn}s to more complex data sets, particularly from critical domains such as medical or scientific data where often relevant concepts are not known in advance, however standard deep learning approaches can learn to focus on confounding factors, \eg for Covid-19 data~\cite{degrave2021ai} or plant phenotyping data~\cite{SchramowskiSSA2020}. We hypothesize the interactive approach of \gls{icsn}s to be beneficial in allowing the machine and human to identify relevant and irrelevant concepts within the data jointly. 
\\
\\
\noindent {\bf Ackowledgements.} The authors thank the anonymous reviewers for their valuable feedback and Cigdem Turan for Figure 1 and 2 sketches. The work has received funding from the BMEL/BLE  under the innovation support program, project ``AuDiSens'' (FKZ28151NA187). It benefited from the Hessian research priority programme LOEWE within the
project WhiteBox as well as from the HMWK cluster project ``The Third Wave of AI.''. 

{\small
\bibliographystyle{ieee_fullname}
\bibliography{egbib}

\begin{thebibliography}{10}\itemsep=-1pt

\bibitem{AlvarezMelisJ18}
David Alvarez{-}Melis and Tommi~S. Jaakkola.
\newblock Towards robust interpretability with self-explaining neural networks.
\newblock In {\em Conference on Neural Information Processing Systems
  (NeurIPS)}, 2018.

\bibitem{archer1966psychological}
E~James Archer.
\newblock The psychological nature of concepts.
\newblock In {\em Analyses of concept learning}, pages 37--49. Elsevier, 1966.

\bibitem{BelemBSB2021}
Catarina Bel{\'e}m, Vladimir Balayan, Pedro Saleiro, and Pedro Bizarro.
\newblock Weakly supervised multi-task learning for concept-based
  explainability.
\newblock {\em arXiv preprint arXiv:2104.12459}, 2021.

\bibitem{BengioCV13}
Yoshua Bengio, Aaron~C. Courville, and Pascal Vincent.
\newblock Representation learning: {A} review and new perspectives.
\newblock {\em IEEE Transactions on Pattern Analysis and Machine Intelligence
  (TPAMI)}, 2013.

\bibitem{BouchacourtTN2018}
Diane Bouchacourt, Ryota Tomioka, and Sebastian Nowozin.
\newblock Multi-level variational autoencoder: Learning disentangled
  representations from grouped observations.
\newblock {\em AAAI Conference on Artificial Intelligence}, 2018.

\bibitem{BowmanIZ2009}
Caitlin~R Bowman, Takako Iwashita, and Dagmar Zeithamova.
\newblock Tracking prototype and exemplar representations in the brain across
  learning.
\newblock {\em eLife}, 2009.

\bibitem{burgess2019monet}
Christopher~P. Burgess, Lo{\"{\i}}c Matthey, Nicholas Watters, Rishabh Kabra,
  Irina Higgins, Matthew Botvinick, and Alexander Lerchner.
\newblock Monet: Unsupervised scene decomposition and representation.
\newblock {\em Computing Research Repository (CoRR)}, 2019.

\bibitem{CarionMSUKZ20}
Nicolas Carion, Francisco Massa, Gabriel Synnaeve, Nicolas Usunier, Alexander
  Kirillov, and Sergey Zagoruyko.
\newblock End-to-end object detection with transformers.
\newblock In {\em European Conference on Computer Vision (ECCV)}, 2020.

\bibitem{CaronMMGBJ20}
Mathilde Caron, Ishan Misra, Julien Mairal, Priya Goyal, Piotr Bojanowski, and
  Armand Joulin.
\newblock Unsupervised learning of visual features by contrasting cluster
  assignments.
\newblock In {\em Conference on Neural Information Processing Systems
  (NeurIPS)}, 2020.

\bibitem{CaronMMG2020}
Mathilde Caron, Ishan Misra, Julien Mairal, Priya Goyal, Piotr Bojanowski, and
  Armand Joulin.
\newblock Unsupervised learning of visual features by contrasting cluster
  assignments.
\newblock In {\em Conference on Neural Information Processing Systems
  (NeurIPS)}, 2020.

\bibitem{ChenLTBR2019}
Chaofan Chen, Oscar Li, Daniel Tao, Alina Barnett, Cynthia Rudin, and
  Jonathan~K Su.
\newblock This looks like that: Deep learning for interpretable image
  recognition.
\newblock In {\em Conference on Neural Information Processing Systems
  (NeurIPS)}, 2019.

\bibitem{ChenKNH2020}
Ting Chen, Simon Kornblith, Mohammad Norouzi, and Geoffrey Hinton.
\newblock A simple framework for contrastive learning of visual
  representations.
\newblock In {\em International Conference on Machine Learning (ICML)}, 2020.

\bibitem{Chen2016}
Xi Chen, Yan Duan, Rein Houthooft, John Schulman, Ilya Sutskever, and Pieter
  Abbeel.
\newblock Infogan: Interpretable representation learning by information
  maximizing generative adversarial nets.
\newblock In {\em Conference on Neural Information Processing Systems
  (NeurIPS)}, 2016.

\bibitem{ChenH2021}
Xinlei Chen and Kaiming He.
\newblock Exploring simple siamese representation learning.
\newblock In {\em IEEE Conference on Computer Vision and Pattern Recognition
  (CVPR)}, 2021.

\bibitem{ChenXH2021}
Xinlei Chen, Saining Xie, and Kaiming He.
\newblock An empirical study of training self-supervised vision transformers.
\newblock {\em arXiv preprint arXiv:2104.02057}, 2021.

\bibitem{CiravegnaBGGL2021}
Gabriele Ciravegna, Pietro Barbiero, Francesco Giannini, Marco Gori, Pietro
  Li{\'o}, Marco Maggini, and Stefano Melacci.
\newblock Logic explained networks.
\newblock {\em arXiv preprint arXiv:2108.05149}, 2021.

\bibitem{ColcombeW2002}
Stanley~J. Colcombe and Robert~S. Wyer.
\newblock The role of prototypes in the mental representation of temporally
  related events.
\newblock {\em Cognitive Psychology}, pages 67--103, 2002.

\bibitem{degrave2021ai}
Alex~J DeGrave, Joseph~D Janizek, and Su-In Lee.
\newblock Ai for radiographic covid-19 detection selects shortcuts over signal.
\newblock {\em Nature Machine Intelligence}, 2021.

\bibitem{Du2020CompositionalVG}
Yilun Du, Shuang Li, and Igor Mordatch.
\newblock Compositional visual generation with energy based models.
\newblock In {\em NeurIPS}, 2020.

\bibitem{du2021unsupervised}
Yilun Du, Shuang Li, Yash Sharma, Josh Tenenbaum, and Igor Mordatch.
\newblock Unsupervised learning of compositional energy concepts.
\newblock {\em NeurIPS}, 2021.

\bibitem{engelckeGenesis}
Martin Engelcke, Adam~R. Kosiorek, Oiwi~Parker Jones, and Ingmar Posner.
\newblock {GENESIS:} generative scene inference and sampling with
  object-centric latent representations.
\newblock In {\em ICLR}, 2020.

\bibitem{eysenck2018fundamentals}
Michael~W Eysenck and Marc Brysbaert.
\newblock {\em Fundamentals of cognition}.
\newblock Routledge, 2018.

\bibitem{fodor1998concepts}
Jerry~A Fodor.
\newblock {\em Concepts: Where cognitive science went wrong}.
\newblock Oxford University Press, 1998.

\bibitem{FrixioneL2012}
Marcello Frixione. and Antonio Lieto.
\newblock Prototypes vs exemplars in concept representation.
\newblock In {\em International Conference on Knowledge Engineering and
  Ontology Development (KEOD)}, 2012.

\bibitem{GhorbaniWZK2019}
Amirata Ghorbani, James Wexler, James~Y Zou, and Been Kim.
\newblock Towards automatic concept-based explanations.
\newblock In {\em Conference on Neural Information Processing Systems
  (NeurIPS)}, 2019.

\bibitem{Girshick15}
Ross~B. Girshick.
\newblock Fast {R-CNN}.
\newblock In {\em International Conference on Computer Vision (ICCV)}, 2015.

\bibitem{GoodfellowPMX2014}
Ian Goodfellow, Jean Pouget-Abadie, Mehdi Mirza, Bing Xu, David Warde-Farley,
  Sherjil Ozair, Aaron Courville, and Yoshua Bengio.
\newblock Generative adversarial nets.
\newblock In {\em Conference on Neural Information Processing Systems
  (NeurIPS)}, 2014.

\bibitem{GreffIODINE}
Klaus Greff, Rapha{\"{e}}l~Lopez Kaufman, Rishabh Kabra, Nick Watters,
  Christopher Burgess, Daniel Zoran, Loic Matthey, Matthew Botvinick, and
  Alexander Lerchner.
\newblock Multi-object representation learning with iterative variational
  inference.
\newblock In {\em International Conference on Machine Learning (ICML)}, 2019.

\bibitem{he2017mask}
Kaiming He, Georgia Gkioxari, Piotr Doll{\'{a}}r, and Ross~B. Girshick.
\newblock Mask {R-CNN}.
\newblock In {\em International Conference on Computer Vision (ICCV)}, 2017.

\bibitem{HeindelFOL2013}
William Heindel, Elena Festa, Brian Ott, Kelly Landy, and David Salmon.
\newblock Prototype learning and dissociable categorization systems in
  alzheimer’s disease.
\newblock {\em Neuropsychologia}, 51, 2013.

\bibitem{HigginsMPBGBML17}
Irina Higgins, Lo{\"{\i}}c Matthey, Arka Pal, Christopher Burgess, Xavier
  Glorot, Matthew Botvinick, Shakir Mohamed, and Alexander Lerchner.
\newblock beta-vae: Learning basic visual concepts with a constrained
  variational framework.
\newblock In {\em International Conference on Learning Representations (ICLR)},
  2017.

\bibitem{Hinton2015}
Geoffrey Hinton, Oriol Vinyals, and Jeff Dean.
\newblock Distilling the knowledge in a neural network.
\newblock {\em arXiv preprint arXiv:1503.02531}, 2015.

\bibitem{Hosoya2018}
Haruo Hosoya.
\newblock Group-based learning of disentangled representations with
  generalizability for novel contents.
\newblock {\em arXiv preprint arXiv:1809.02383}, 2018.

\bibitem{Hosoya2019}
Haruo Hosoya.
\newblock Group-based learning of disentangled representations with
  generalizability for novel contents.
\newblock In {\em Joint Conference on Artificial Intelligence (IJCAI)}, 2019.

\bibitem{Hosoya19}
Haruo Hosoya.
\newblock Group-based learning of disentangled representations with
  generalizability for novel contents.
\newblock In Sarit Kraus, editor, {\em Joint Conference on Artificial
  Intelligence (IJCAI)}, 2019.

\bibitem{BillardDPM2018}
Yordan Hristov, Alex Lascarides, and Subramanian Ramamoorthy.
\newblock Interpretable latent spaces for learning from demonstration.
\newblock In {\em Conference on Robot Learning (CoRL)}, 2018.

\bibitem{HuangZWY2021}
Shixin Huang, Xiangping Zeng, Si Wu, Zhiwen Yu, Mohamed Azzam, and Hau-San
  Wong.
\newblock Behavior regularized prototypical networks for semi-supervised
  few-shot image classification.
\newblock {\em Pattern Recognition}, 2021.

\bibitem{IbbotsonT2009}
Paul Ibbotson and Michael Tomasello.
\newblock Prototype constructions in early language acquisition.
\newblock {\em Camebridge University Press}, pages 59--85, 2009.

\bibitem{JaekelSW2008}
Frank J{\"a}kel, Bernhard Sch{\"o}lkopf, and Felix~A. Wichmann.
\newblock Generalization and similarity in exemplar models of categorization:
  Insights from machine learning.
\newblock {\em Psychonomic Bulletin {\&} Review}, 2008.

\bibitem{JangGP17}
Eric Jang, Shixiang Gu, and Ben Poole.
\newblock Categorical reparameterization with gumbel-softmax.
\newblock In {\em International Conference on Learning Representations (ICLR)},
  2017.

\bibitem{InsuWMG2021}
Insu Jeon, Wonkwang Lee, Myeongjang Pyeon, and Gunhee Kim.
\newblock Ib-gan: Disengangled representation learning with information
  bottleneck generative adversarial networks.
\newblock In {\em AAAI Conference on Artificial Intelligence}, 2021.

\bibitem{KimWGCW2018}
Been Kim, Martin Wattenberg, Justin Gilmer, Carrie Cai, James Wexler, Fernanda
  Viegas, and Rory sayres.
\newblock Interpretability beyond feature attribution: Quantitative testing
  with concept activation vectors ({TCAV}).
\newblock In {\em Proceedings of the 35th International Conference on Machine
  Learning}, Proceedings of Machine Learning Research (PMLR), 2018.

\bibitem{KimM18}
Hyunjik Kim and Andriy Mnih.
\newblock Disentangling by factorising.
\newblock In {\em International Conference on Machine Learning (ICML)}, 2018.

\bibitem{KohNTMPKL20}
Pang~Wei Koh, Thao Nguyen, Yew~Siang Tang, Stephen Mussmann, Emma Pierson, Been
  Kim, and Percy Liang.
\newblock Concept bottleneck models.
\newblock In {\em Proceedings of Machine Learning Research (PMLR)}, 2020.

\bibitem{LiLCR2018}
Oscar Li, Hao Liu, Chaofan Chen, and Cynthia Rudin.
\newblock Deep learning for case-based reasoning through prototypes: A neural
  network that explains its predictions.
\newblock In {\em AAAI Conference on Artificial Intelligence}, 2018.

\bibitem{LiCWW2019}
Xiaoqiang Li, Liangbo Chen, Lu Wang, Pin Wu, and Weiqin Tong.
\newblock Scgan: Disentangled representation learning by adding similarity
  constraint on generative adversarial nets.
\newblock {\em IEEE Access}, 2019.

\bibitem{ZhizhongDBJ2016}
Zhizhong Li and Derek Hoiem.
\newblock Learning without forgetting.
\newblock In Bastian Leibe, Jiri Matas, Nicu Sebe, and Max Welling, editors,
  {\em European Conference on Computer Vision (ECCV)}, 2016.

\bibitem{LinSPACE}
Zhixuan Lin, Yi{-}Fu Wu, Skand~Vishwanath Peri, Weihao Sun, Gautam Singh, Fei
  Deng, Jindong Jiang, and Sungjin Ahn.
\newblock {SPACE:} unsupervised object-oriented scene representation via
  spatial attention and decomposition.
\newblock In {\em International Conference on Learning Representations (ICLR)},
  2020.

\bibitem{LocatelloBLRG2019}
Francesco Locatello, Stefan Bauer, Mario Lucic, Gunnar Raetsch, Sylvain Gelly,
  Bernhard Sch{\"o}lkopf, and Olivier Bachem.
\newblock Challenging common assumptions in the unsupervised learning of
  disentangled representations.
\newblock In {\em International Conference on Machine Learning (ICML)}, 2019.

\bibitem{LocatelloBLRGSB19}
Francesco Locatello, Stefan Bauer, Mario Lucic, Gunnar R{\"{a}}tsch, Sylvain
  Gelly, Bernhard Sch{\"{o}}lkopf, and Olivier Bachem.
\newblock Challenging common assumptions in the unsupervised learning of
  disentangled representations.
\newblock In {\em International Conference on Machine Learning (ICML)}, 2019.

\bibitem{LocatelloPRSBT20}
Francesco Locatello, Ben Poole, Gunnar R{\"{a}}tsch, Bernhard Sch{\"{o}}lkopf,
  Olivier Bachem, and Michael Tschannen.
\newblock Weakly-supervised disentanglement without compromises.
\newblock In {\em International Conference on Machine Learning (ICML)}, 2020.

\bibitem{SlotAttention}
Francesco Locatello, Dirk Weissenborn, Thomas Unterthiner, Aravindh Mahendran,
  Georg Heigold, Jakob Uszkoreit, Alexey Dosovitskiy, and Thomas Kipf.
\newblock Object-centric learning with slot attention.
\newblock In {\em Conference on Neural Information Processing Systems
  (NeurIPS)}, 2020.

\bibitem{MaddisonMT17}
Chris~J. Maddison, Andriy Mnih, and Yee~Whye Teh.
\newblock The concrete distribution: {A} continuous relaxation of discrete
  random variables.
\newblock In {\em International Conference on Learning Representations (ICLR)},
  2017.

\bibitem{MarcosFLF2021}
Diego Marcos, Ruth Fong, Sylvain Lobry, R{\'e}mi Flamary, Nicolas Courty, and
  Devis Tuia.
\newblock Contextual semantic interpretability.
\newblock In Hiroshi Ishikawa, Cheng-Lin Liu, Tomas Pajdla, and Jianbo Shi,
  editors, {\em Asia Conference on Computer Vision (ACCV)}, 2021.

\bibitem{McCloskeyN1989}
Michael McCloskey and Neal~J. Cohen.
\newblock Catastrophic interference in connectionist networks: The sequential
  learning problem.
\newblock In {\em Psychology of learning and motivation}, volume~24, pages
  109--165. Academic Press, 1989.

\bibitem{MemmelGM2021}
Marius Memmel, Camila Gonzalez, and Anirban Mukhopadhyay.
\newblock Adversarial continual learning for multi-domain hippocampal
  segmentation.
\newblock In {\em Domain Adaptation and Representation Transfer (DART) at
  International Conference on Medical Image Computing and Computer Assisted
  Intervention (MICCAI)}, 2021.

\bibitem{ZMichieliZ21}
Umberto Michieli and Mete Ozay.
\newblock Prototype guided federated learning of visual feature
  representations.
\newblock {\em arXiv preprint arXiv:2105.08982}, 2021.

\bibitem{MeilaZ2021}
Graziano Mita, Maurizio Filippone, and Pietro Michiardi.
\newblock An identifiable double vae for disentangled representations.
\newblock In {\em International Conference on Machine Learning (ICML)}, 2021.

\bibitem{NguyenRMY2020}
Thu~H Nguyen-Phuoc, Christian Richardt, Long Mai, Yongliang Yang, and Niloy
  Mitra.
\newblock Blockgan: Learning 3d object-aware scene representations from
  unlabelled images.
\newblock In {\em Conference on Neural Information Processing Systems
  (NeurIPS)}, 2020.

\bibitem{NiemeyerG2020}
Michael Niemeyer and Andreas Geiger.
\newblock Giraffe: Representing scenes as compositional generative neural
  feature fields.
\newblock In {\em IEEE Conference on Computer Vision and Pattern Recognition
  (CVPR)}, 2021.

\bibitem{OppenheierTK2013}
Daniel~M. Oppenheimer, Joshua~B. Tenenbaum, and Tevye~R. Krynski.
\newblock Chapter six - categorization as causal explanation: Discounting and
  augmenting in a bayesian framework.
\newblock In {\em Categorization as causal reasoning}, Psychology of Learning
  and Motivation, pages 203--231. Academic Press, 2013.

\bibitem{PahdePKN21}
Frederik Pahde, Mihai Puscas, Tassilo Klein, and Moin Nabi.
\newblock Multimodal prototypical networks for few-shot learning.
\newblock In {\em IEEE Winter Conference on Applications of Computer Vision
  (WACV)}, 2021.

\bibitem{PatacchiolaS2020}
Massimiliano Patacchiola and Amos Storkey.
\newblock Self-supervised relational reasoning for representation learning.
\newblock {\em arXiv preprint arXiv:2006.05849}, 2020.

\bibitem{PrabhudesaiLPTH21}
Mihir Prabhudesai, Shamit Lal, Darshan Patil, Hsiao-Yu Tung, Adam~W Harley, and
  Katerina Fragkiadaki.
\newblock Disentangling 3d prototypical networks for few-shot concept learning.
\newblock In {\em International Conference on Learning Representations (ICLR)},
  2021.

\bibitem{Ratcliff1990}
Roger Ratcliff.
\newblock Connectionist models of recognition memory: constraints imposed by
  learning and forgetting functions.
\newblock {\em Psychological review}, 97(2):285, 1990.

\bibitem{RedmonDGF16}
Joseph Redmon, Santosh~Kumar Divvala, Ross~B. Girshick, and Ali Farhadi.
\newblock You only look once: Unified, real-time object detection.
\newblock In {\em IEEE Conference on Computer Vision and Pattern Recognition
  (CVPR)}, 2016.

\bibitem{Robins1993}
A. Robins.
\newblock Catastrophic forgetting in neural networks: the role of rehearsal
  mechanisms.
\newblock In {\em Proceedings 1993 The First New Zealand International
  Two-Stream Conference on Artificial Neural Networks and Expert Systems},
  pages 65--68, 1993.

\bibitem{RossCHGD21}
Andrew~Slavin Ross, Nina Chen, Elisa~Zhao Hang, Elena~L. Glassman, and Finale
  Doshi{-}Velez.
\newblock Evaluating the interpretability of generative models by interactive
  reconstruction.
\newblock In {\em Conference on Human Factors in Computing System (CHI)}, 2021.

\bibitem{Ross2017right}
Andrew~Slavin Ross, Michael~C. Hughes, and Finale Doshi-Velez.
\newblock Right for the right reasons: Training differentiable models by
  constraining their explanations.
\newblock In {\em Joint Conference on Artificial Intelligence (IJCAI)}, 2017.

\bibitem{SchramowskiSSA2020}
Patrick Schramowski, Wolfgang Stammer, Stefano Teso, Anna Brugger, Franziska
  Herbert, Xiaoting Shao, Hans-Georg Luigs, Anne-Katrin Mahlein, and Kristian
  Kersting.
\newblock Making deep neural networks right for the right scientific reasons by
  interacting with their explanations.
\newblock {\em Nature Machine Intelligence}, 2020.

\bibitem{Seel2012}
Norbert~M. Seel, editor.
\newblock {\em Prototype}, pages 2714--2714.
\newblock Springer US, 2012.

\bibitem{Selvaraju2019taking}
Ramprasaath~Ramasamy Selvaraju, Stefan Lee, Yilin Shen, Hongxia Jin, Shalini
  Ghosh, Larry~P. Heck, Dhruv Batra, and Devi Parikh.
\newblock Taking a {HINT:} leveraging explanations to make vision and language
  models more grounded.
\newblock In {\em International Conference on Computer Vision (ICCV)}, 2019.

\bibitem{Senior0JKSGQZNB20}
Andrew~W. Senior, Richard Evans, John Jumper, James Kirkpatrick, Laurent Sifre,
  Tim Green, Chongli Qin, Augustin Z{\'{\i}}dek, Alexander W.~R. Nelson, Alex
  Bridgland, Hugo Penedones, Stig Petersen, Karen Simonyan, Steve Crossan,
  Pushmeet Kohli, David~T. Jones, David Silver, Koray Kavukcuoglu, and Demis
  Hassabis.
\newblock Improved protein structure prediction using potentials from deep
  learning.
\newblock {\em Nat.}, 577(7792):706--710, 2020.

\bibitem{ShuCKEP20}
Rui Shu, Yining Chen, Abhishek Kumar, Stefano Ermon, and Ben Poole.
\newblock Weakly supervised disentanglement with guarantees.
\newblock In {\em International Conference on Learning Representations (ICLR)},
  2020.

\bibitem{Smith2014}
J.~David Smith.
\newblock Prototypes, exemplars, and the natural history of categorization.
\newblock {\em Psychonomic bulletin {\&} review}, 2014.

\bibitem{SnellSZ17}
Jake Snell, Kevin Swersky, and Richard Zemel.
\newblock Prototypical networks for few-shot learning.
\newblock In {\em Conference on Neural Information Processing Systems
  (NeurIPS)}, 2017.

\bibitem{StammerSK2021}
Wolfgang Stammer, Patrick Schramowski, and Kristian Kersting.
\newblock Right for the right concept: Revising neuro-symbolic concepts by
  interacting with their explanations.
\newblock In {\em IEEE Conference on Computer Vision and Pattern Recognition
  (CVPR)}, 2021.

\bibitem{Taylor2003}
J.R. Taylor.
\newblock {\em Linguistic Categorization}.
\newblock OUP Oxford, 2003.

\bibitem{TesoK2019}
Stefano Teso and Kristian Kersting.
\newblock Explanatory interactive machine learning.
\newblock In {\em AAAI/ACM Conference on AI, Ethics, and Society (AIES)}, 2019.

\bibitem{TschannenBL2018}
Michael Tschannen, Olivier Bachem, and Mario Lucic.
\newblock Recent advances in autoencoder-based representation learning.
\newblock {\em arXiv preprint arXiv:1812.05069}, 2018.

\bibitem{VCB2020}
Matthew~J. Vowels, Necati~Cihan Camgoz, and Richard Bowden.
\newblock Gated variational autoencoders: Incorporating weak supervision to
  encourage disentanglement.
\newblock In {\em IEEE International Conference on Automatic Face and Gesture
  Recognition (FG)}, 2020.

\bibitem{WuH18}
Yuxin Wu and Kaiming He.
\newblock Group normalization.
\newblock In {\em European Conference on Computer Vision (ECCV)}, 2018.

\bibitem{YangLCS2021}
Mengyue Yang, Furui Liu, Zhitang Chen, Xinwei Shen, Jianye Hao, and Jun Wang.
\newblock Causalvae: Disentangled representation learning via neural structural
  causal models.
\newblock In {\em IEEE Conference on Computer Vision and Pattern Recognition
  (CVPR)}, 2021.

\bibitem{YehKAL2020}
Chih-Kuan Yeh, Been Kim, Sercan Arik, Chun-Liang Li, Tomas Pfister, and Pradeep
  Ravikumar.
\newblock On completeness-aware concept-based explanations in deep neural
  networks.
\newblock In {\em Conference on Neural Information Processing Systems
  (NeurIPS)}, 2020.

\bibitem{yi2018neural}
Kexin Yi, Jiajun Wu, Chuang Gan, Antonio Torralba, Pushmeet Kohli, and Josh
  Tenenbaum.
\newblock Neural-symbolic {VQA:} disentangling reasoning from vision and
  language understanding.
\newblock In {\em Conference on Neural Information Processing Systems
  (NeurIPS)}, 2018.

\bibitem{ZaidiBGC2020}
Julian Zaidi, Jonathan Boilard, Ghyslain Gagnon, and Marc-Andr{\'e} Carbonneau.
\newblock Measuring disentanglement: A review of metrics.
\newblock {\em arXiv preprint arXiv:2012.09276}, 2020.

\bibitem{Zeithamova2012}
Dagmar Zeithamova.
\newblock {\em Prototype Learning Systems}, pages 2715--2718.
\newblock Springer US, 2012.

\bibitem{ZhuXT2021}
Xinqi Zhu, Chang Xu, and Dacheng Tao.
\newblock Where and what? examining interpretable disentangled representations.
\newblock In {\em IEEE Conference on Computer Vision and Pattern Recognition
  (CVPR)}, 2021.

\end{thebibliography}
}

\clearpage

\onecolumn







\begin{center}
\section*{Interactive Disentanglement: \\Learning Concepts by Interacting with their Prototype Representations\\ \vspace{1em} Supplementary Materials}
\end{center}


\maketitle

\thispagestyle{empty} 

\subsection*{Hyperparameters and model details}


\subsubsection*{Interactive Concept Swapping Networks}

In our experiments, the prototype slots were initialized randomly from a truncated Gaussian distribution with mean $\mu = 0$, variance $\sigma^2=0.5$, minimum $a=-1$, and maximum $b=1$. The encoder used in our experiments was a convolutional neural network with residual connections and ReLU activations. Each read-out encoder is a linear layer with LeakyReLU activations. Lastly, the decoder architecture was again a neural network with transposed convolutions and also here residual layers.

\noindent In the standard \acrshort{ecr} experiments with \acrshort{icsn}s, $J=3$, $Z=512$, $Q=128$, $K=6$ for each $j \in [1, ..., J]$, $N=128$ and $\tau$ was decreased every 1000 epochs with steps $[2., 0.5, 0.1, 0.01, 0.001, 0.0001, 0.00001]$ over 8000 epochs in total.

\noindent Notably, group normalization as proposed by Wu \textit{et al.}~\cite{WuH18} was applied after extracting the concept encodings via the read-out encoders (performed in {\tt collectedReadOutEncoders} in Alg.~\ref{alg:icsn}).

\noindent Pseudo code can be found in Alg.~\ref{alg:icsn} and Alg.~\ref{alg:icsn2}.

\subsubsection*{Baseline models}
For the experiments with Cat-VAE, the softmax temperature was set to $\tau=0.1$ and each categorical distribution had $k=6$ categories. The number of latent variables was set to $3$ for Cat-VAE, $\beta$-VAE, Ada-VAE, and VAE runs. For all baselines, the encoder and decoder consisted of a convolutional and transposed convolutional network. In all experiments, $\beta = 4.$, except for Ada-VAE, where $\beta = 1.$, as recommended by Locatello \textit{et al.}~\cite{LocatelloPRSBT20}. All baseline models were trained for 2000 epochs.

\subsubsection*{Linear probing}
The linear models for probing the latent representations of the different model configurations were a decision tree and logistic regression model. The max depth of the decision tree was set to 8. The logistic regression model was run with parameters $C=0.316$ and the maximum number of iterations at 1000. Both the decision tree and logistic regression model were trained with a fixed random seed.

\subsection*{Details on simulated interactions}

The simulated user interactions were performed via an $L_2$ regulatory loss term on the latent codes $y$. 

\noindent In case a user tells an iCSN \textit{not} to use a specific prototype slot of the superordinate concept $j$ and slot $k$, this loss corresponds to: $MSE(y_{j\cdot k + k}, \mathbf{0})$, where $y_{j\cdot k + k}$ corresponds to the value of $y$ at position $j\cdot k + k$ and $\mathbf{0}$ being a vector of length $N$.

\noindent When a user provides a subset of examples with corresponding desired prototype slot IDs the loss term corresponds to: $MSE(y^{subset}_{j\cdot k + k}, \mathbf{1})$ with $\mathbf{1}$ of length equal to the number of samples in \textit{subset}. The subset of examples in our simulated interactions were identified via the ground truth labels, e.g., for identifying the subset of images containing a pentagon. The interactions for learning a novel basic concept followed the same procedure.

\noindent To allow the model to update is latent space via interactions we increased $\tau=0.00001$ back to $\tau=0.0001$.

\begin{algorithm}[t]
    \SetKwFunction{collectedReadOutEncoders}{collectedReadOutEncoders}
    \SetKwFunction{computeProtoDistance}{computeProtoDistance}
    \SetKwInOut{KwIn}{Input}
    \SetKwInOut{KwOut}{Output}
    
    \KwIn{Image pair $x\in\mathbb{R}^{D}$, $x'\in\mathbb{R}^{D}$, known share IDs $\mathrm{v}$.}
    \KwOut{Image reconstructions $\hat{x}\in\mathbb{R}^{D}$, $\hat{x}'\in\mathbb{R}^{D}$, and latent codes $y \in [0, 1]^{J\cdot K}$, $y' \in [0, 1]^{J\cdot K}$}

    \tcp{Forward pass through initial encoder. $z\in \mathbb{R}^{Z}$}
    $z \leftarrow f(x)$\\
    $z' \leftarrow f(x')$\\

    \tcp{Forward pass through $J$ read-out encoders. $\phi\in \mathbb{R}^{J \times Q}$}
    $\phi \leftarrow \collectedReadOutEncoders(z)$\\
    $\phi' \leftarrow \collectedReadOutEncoders(z')$\\

    \tcp{Compute the distance of each concept encoding to all prototype slots of its corresponding category $j$.}
    $y \leftarrow \computeProtoDistance(\phi, \mathrm{v})$\\
    $y' \leftarrow \computeProtoDistance(\phi', \mathrm{v})$\\

    \tcp{Reconstruct the images from the prototype distance codes.}
    $\hat{y} \leftarrow g(y)$\\
    $\hat{y'} \leftarrow g(y')$\\

    \caption{Interactive Concept Swapping Network -- pair images forward pass \label{alg:icsn}}
\end{algorithm}

\begin{algorithm}[t]
    \SetKwFunction{softmaxDotProduct}{softmaxDotProduct}
    \SetKwFunction{softmaxNormTau}{softmaxNormTau}
    \SetKwInOut{KwIn}{Input}
    \SetKwInOut{KwOut}{Output}
    \SetKwInOut{Given}{Given}
    
    \KwIn{Concept encodings $\phi\in\mathbb{R}^{J \times Q}$, $\phi' \in\mathbb{R}^{J \times Q}$}
    \Given{Set of prototype slot codebooks $\Theta := [\mathrm{P}_1, ..., \mathrm{P}_J] \in \mathbb{R}^{J \times Q \times K}$, softmax temperature $\tau$, and share IDs $\mathrm{v}$.}
    \KwOut{Latent codes $y \in [0, 1]^{J\cdot K}$, $y' \in [0, 1]^{J\cdot K}$}

    \tcp{For every superordinate concept category}
     \For{$j \leftarrow 0$ \KwTo $J-1$}{
        \tcp{Dot-product between concept encoding and all prototype slots from codebook $\mathrm{P}_j$.}
        $s_j \leftarrow \softmaxDotProduct(\phi_j, \mathrm{P}_j)$\\
        $s'_j \leftarrow \softmaxDotProduct(\phi'_j, \mathrm{P}_j)$\\
        
        \tcp{Compute normalizing weighted softmax.}
        $\Pi_j \leftarrow \softmaxNormTau(s_j, \tau)$\\
        $\Pi'_j \leftarrow \softmaxNormTau(s'_j, \tau)$
        
    }

    \tcp{Swap the distance codes at the position corresponding to the shared IDs.}
    $y \leftarrow [\Pi_{1}, ..., \mathbf{\Pi'_{v}}, ..., \Pi_{J}]$\\
    $y' \leftarrow [\Pi'_{1}, ..., \mathbf{\Pi_{v}}, ..., \Pi'_{J}]$

    \caption{computeProtoDistance \label{alg:icsn2}}
\end{algorithm}

\clearpage 


\end{document}